SpaceOps-2025, ID # 140

# On the Role of AI in Managing Satellite Constellations: Insights from the ConstellAI Project

Gregory F. Stock[a], Juan A. Fraire[ab*], Holger Hermanns[a], Jędrzej Mosiężny[c], Yusra Al-Khazraji[c],
Julio Ramírez Molina[c], Evridiki V. Ntagiou[d]

[a] *Saarland University – Computer Science, Saarland Informatics Campus, 66123 Saarbrücken, Germany*
[b] *Inria, INSA Lyon, CITI, UR3720, 69621 Villeurbanne, France*
[c] *GMV GmbH, 64293 Darmstadt, Germany*
[d] *European Space Agency (ESA-ESOC), 64293 Darmstadt, Germany*
\* Corresponding author

**Abstract**

The rapid expansion of satellite constellations in near-Earth orbits presents significant challenges in satellite network management, requiring innovative approaches for efficient, scalable, and resilient operations. This paper explores the role of Artificial Intelligence (AI) in optimizing the operation of satellite mega-constellations, drawing from the ConstellAI project funded by the European Space Agency (ESA). A consortium comprising GMV GmbH, Saarland University, and Thales Alenia Space collaborates to develop AI-driven algorithms and demonstrates their effectiveness over traditional methods for two crucial operational challenges: data routing and resource allocation. In the routing use case, Reinforcement Learning (RL) is used to improve the end-to-end latency by learning from historical queuing latency, outperforming classical shortest path algorithms. For resource allocation, RL optimizes the scheduling of tasks across constellations, focussing on efficiently using limited resources such as battery and memory.

Both use cases were tested for multiple satellite constellation configurations and operational scenarios, resembling the real-life spacecraft operations of communications and Earth observation satellites. This research demonstrates that RL not only competes with classical approaches but also offers enhanced flexibility, scalability, and generalizability in decision-making processes, which is crucial for the autonomous and intelligent management of satellite fleets. The findings of this activity suggest that AI can fundamentally alter the landscape of satellite constellation management by providing more adaptive, robust, and cost-effective solutions.

**Acronyms/Abbreviations**

**AI** Artificial Intelligence
**AT** Acquisition Target
**EO** Earth Observation
**ESA** European Space Agency
**ESOC** European Space Operations Centre
**GS** Ground Station
**ISL** Inter-Satellite Link
**LEO** Low-Earth Orbit
**NN** Neural Network
**PPO** Proximal Policy Optimization
**RL** Reinforcement Learning
**SA** Simulated Annealing
**SatCom** Satellite Communication

## 1. Introduction

The space industry is undergoing a significant transformation, marked by the rapid growth of satellite (mega-)constellations and the emergence of the "New Space" paradigm. This new era is characterized by increased commercial space ventures and a shift towards more innovative and cost-effective approaches to space operations. As satellite fleets continue to expand, traditional management methods face substantial challenges in handling the complexity and scale of these large constellations.

Artificial Intelligence (AI) has emerged as a critical tool in addressing these challenges. The scalability issues inherent in managing extensive satellite networks necessitate the adoption of AI technologies. AI has the potential to revolutionize satellite operations by providing advanced solutions that traditional optimization techniques cannot achieve. These conventional methods often fall short in handling the increasing level of complexity and dynamic nature of real-world operations, making AI an increasingly preferred choice for effective fleet management.

The ConstellAI project, officially titled "Artificial Intelligence for Large Fleet Network Management," aims to leverage AI to enhance the management of large satellite fleets. Funded by the European Space Operations Centre (ESOC) and the European Space Agency (ESA), this project brings together a consortium of academic and industry partners, including GMV GmbH (Germany), Saarland University (Germany), and Thales Alenia Space (France). Additionally, the project benefits from the expertise of the two satellite operators Eutelsat (France) and Planet (Germany), who serve as consultants for Satellite Communication (SatCom) and Earth Observation (EO), respectively. Through these strategic partnerships, ConstellAI seeks to develop innovative AI-driven solutions to meet the evolving demands of satellite fleet management. In the context of the project, we explored two crucial operational challenges: data routing and resource allocation across constellations, for which we developed AI solutions.

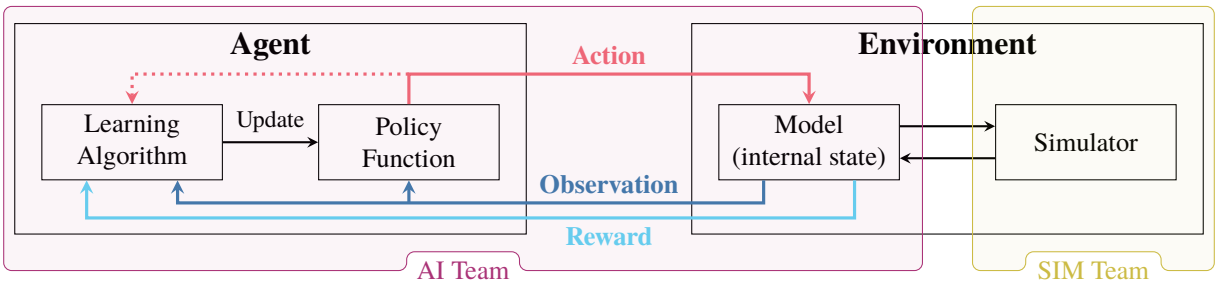

Fig. 1: Reinforcement Learning framework showing the interaction between the AI agent and environment.

This paper builds upon preliminary results presented in [1] and provides the following novel contributions:

- A complete and formal definition of the AI models used in both routing and resource allocation use cases, including the state/action/reward structure and learning strategy.

- A detailed performance comparison of AI-based techniques against classical baselines through extensive simulations.

- A comprehensive flexibility analysis under failure scenarios to assess robustness.

- A discussion of the trade-offs in execution time, stability, and adaptability for each AI technique.

The remainder of this paper is structured as follows: Section 2 introduces the background of satellite operations and motivates the use of AI. Next, we introduce the two use cases in Section 3 (data routing) and Section 4 (resource allocation). In Section 5, we present comprehensive evaluation results from our verification and validation analysis. Section 6 describes the technical details of our tool infrastructure and Section 7 concludes the paper.

## 2. Background

Satellite operations have traditionally relied on well-established methods of planning and scheduling, often based on deterministic algorithms and rule-based heuristics. Classical approaches involve ground-based mission control centres that monitor, schedule, and command satellites to execute pre-defined tasks. For decades, these methods have been effective for single-satellite missions and small constellations. However, the rapid deployment of large satellite constellations, particularly in Low-Earth Orbit (LEO), is creating new challenges in network management. As the number of assets in orbit increases, so does the complexity of managing their operations, making manual and semi-automated approaches to satellite operations increasingly insufficient [2, 3]. These limitations of traditional approaches highlight the necessity for advanced and innovative optimization techniques such as AI to improve operational efficiency [4, 5].

Two key use cases for satellite constellations are Earth Observation (EO) and Satellite Communication (SatCom) missions. EO satellites are essential for monitoring environmental changes, assessing the impact of disasters, and conducting maritime surveillance. For example, climate change monitoring relies on satellite-based remote sensing to track temperature variations, rising sea levels, and deforestation. Notable missions are Sentinel-3 [6, 7, 8] and Sentinel-6 [9, 10, 11] from ESA's Copernicus programme [12] as well as NASA's Landsat satellites [13, 14]. Weather forecasting benefits from high-resolution imagery and atmospheric data collected by EO satellites (e.g. Sentinel-5P [15, 16]). In disaster management, satellites play a critical role in detecting and assessing natural disasters such as floods and wildfires, providing near real-time data to enable rapid response. For example, Planet's Dove and SkySat satellites provide high-quality satellite images for commercial and research applications [17, 18]. Maritime surveillance is another critical application where satellites (e.g. Sentinel-1 [19, 20], HawkEye 360 [21]) help track illegal fishing activities and monitor ship movements in remote oceanic regions. Lastly, ESA's PhiSat mission integrates AI onboard the satellite to enhance image processing and data filtering before transmission to the ground, reducing bandwidth consumption and improving efficiency [22].

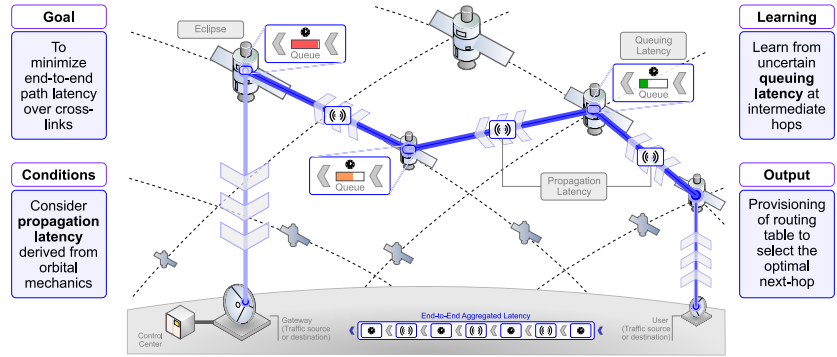

Fig. 2: Overview of Use Case 1: Routing.

SatCom missions, on the other hand, aim to provide global connectivity and secure communication services. For example, ESA's IRIS² initiative is designed to offer secure and resilient communication for European governments and businesses [23]. Constellations such as SpaceX's Starlink [24] and Eutelsat's OneWeb [25] offer low-latency broadband Internet services to enhance global Internet access, particularly in underserved areas. Satellite telephony/data services rely on constellations like Iridium (Iridium Communications) [26] and O3b mPOWER (SES) [27] to provide connectivity or emergency response in remote locations where terrestrial networks are unavailable (e.g. while hiking or in an aircraft). Other major players include the operators Intelsat, Viasat, Inmarsat, and Hispasat, all of which contribute to the global satellite communication infrastructure.

Both types of missions—EO and SatCom—require continuous optimization to ensure seamless connectivity and efficient use of resources. Therefore, we have selected one example use case from each category for which we are developing AI solutions in the ConstellAI project.

AI, in particular Reinforcement Learning (RL), has emerged as a promising technology towards automating and optimizing complex tasks. In the context of satellite operations, AI techniques enable automated decision-making and reduce human intervention, improving overall mission efficiency [4, 5]. RL is a subset of machine learning in which an agent interacts with an environment and learns from the results of its actions to refine its policy over time to achieve optimal results [28]. The typical RL workflow is shown in Fig. 1: An AI agent interacts with a (simulated) satellite environment that models the dynamics of the satellite network. The agent performs actions according to its internal policy and observes the changed state, along with a reward value that gives the agent feedback on the quality of the selected action. The agent continuously updates its policy to improve the decision-making and eventually converge to the optimal strategy. Within ConstellAI, the *AI Team* was responsible to develop and implement the AI agent as well as its interaction with the environment while another group, the *SIM Team*, developed a simulator that interfaces with the environment. In principle, the simulator is a stand-alone component that generates the necessary data to train the AI and assess its solutions, e.g. propagating the satellite orbits or computing access intervals. At its core, it uses GODOT, the ESA/ESOC flight dynamics software for performing orbit-related computations [29].

## 3. Use Case 1: Routing

The first use case addresses packet routing in SatCom constellations, where data packets traverse a network of interconnected satellites. The primary objective is to minimize the end-to-end latency throughout the satellite network by considering both the propagation delays between satellites and further uncertainties such as queuing latencies. Fig. 2 provides a high-level overview of the use case, highlighting how packets are routed from the user terminal to the ground station gateway.

Using RL algorithms, an agent learns from historical queueing latency data to dynamically update a Q-table. This Q-table indicates the most optimal next-hop nodes based on current network conditions and can be easily cast into the routing tables on each satellite. This way, our method continuously adapts the routing strategy to minimize the total end-to-end route delay, ensuring efficient data flow across a constantly changing satellite network. In contrast to AI solutions, classical optimization algorithms such as Dijkstra's shortest path can typically only minimize the propagation delay since the current queueing delays are unknown. Note that, in practice, the route with minimal propagation delay is not necessarily also the one that minimizes the overall route delay [30].

### 3.1 Model Definition

In this section, we give the formal definition of our AI satellite network environment as well as the mathematical details of the Q-learning algorithm. Our environment considers an interconnected satellite network with known orbits, which allows us to pre-compute the propagation latencies between adjacent satellites.

Formally, we model a satellite network as a graph $\mathcal{G} = (V, E)$, where $V = \{s_1, s_2, \ldots, s_n\}$ represents the set of vertices (i.e. satellites), and $E \subseteq V \times V$ represents the set of edges (i.e. inter-satellite links (ISLs)) between nodes in the graph. An edge $(u, v) \in E$ indicates a direct communication link between the two satellites $u$ and $v$. The total number of nodes $n = |V|$ denotes the size of the satellite network. Each satellite is identified by a unique index (a satellite ID in $\{1, \ldots, n\}$) or, alternatively, by a coordinate tuple $(\Omega, \nu)$ in a grid-based topology, where $\Omega$ is the longitude of the ascending node and $\nu$ the true anomaly (or argument of latitude for circular orbits). While the implementation mostly works with the index internally, the coordinate tuple is helpful when visualizing the satellites in the graphical user interface.

In the following, we assume a satellite network $\mathcal{G} = (V, E)$ as well as a specific source node $src \in V$ and destination node $dst \in V$. Using RL, the goal of the AI model is to efficiently route data packets from $src$ through the network to $dst$. Efficiently means that the model should learn from historical data on queueing latency to find the route that minimizes the total end-to-end route delay.

**State Space.** The state space of our RL model is mathematically defined as $S = V = \{s_1, s_2, \ldots, s_n\}$. A state $s_i \in S$ indicates that the data packet is currently at the $i$-th satellite in the network.

**Action Space.** Actions correspond to forwarding the data packet from the current satellite to a specific destination node. The action space $A = \{a_1, a_2, \ldots, a_n\}$ therefore includes an action for each satellite, where action $a_i$ means to send the data packet (from the current satellite) to $s_i$. The action space $A$ and state space $S$ are therefore isomorphic $A \cong S$ (i.e. there exists a bijective mapping between $A$ and $S$). Hence, we will use an action $a_i$ and its successor satellite $s_i$ interchangeably in the following.

Since not all satellites are connected to each other, the action space for each satellite is restricted to only the feasible actions. We define the (feasible) action space $A^f(s)$ for a satellite node $s$ to include all possible transitions to adjacent satellites, i.e. $A^f(s) = \{a_i \in A \mid (s, s_i) \in E\}$.

**Policy.** The routing policy determines the next-hop decision (i.e. to which neighbour the packet should be forwarded) at each state $s$ based on Q-values that are stored in a Q-table $Q \in \mathbb{R}^{n \times n}$. This table maps state-action pairs to a number indicating how "good" an action is considered in a given state. We write $Q(s, a)$ to denote the entry in the Q-table for state $s$ and action $a$. In the beginning, all entries are initialized to zero.

The goal of Q-routing is to continuously adapt the Q-table until the values converge so that the chosen actions minimize the overall time for packet delivery across the network. At each state $s$, the (currently) best action $a^\star$ is chosen as the action $a^\star \coloneqq \arg\max_{a \in A^f(s)} Q(s, a)$ with the highest Q-value. At the end, the final route can be reconstructed by starting at the source node and following the best actions until the destination is reached—while keeping track of visited nodes to avoid loops. In case the best action would create a loop, the second-best action is used instead and so forth.

**Reward.** The reward $R(s, a)$ for choosing action $a$ in state $s$ reflects the success and quality of this routing decision and serves as an important feedback to guide the agent's learning process. The goal is to minimize the end-to-end routing delay by discouraging high-latency paths. Since actions with higher Q-values are preferred, we define the reward as a negative function of the total latency so that lower latencies cause higher Q-values:

$$R(s, a) \coloneqq -(L_{prop}(s, a) + L_{queue}(s, a) + L_{target}(a)) \tag{1}$$

In RL, rewards are often negative to represent costs, penalties, or other undesirable outcomes that an agent should learn to avoid. Here, $L_{prop}(s, a)$ is the (static) propagation latency from satellite $s$ to its neighbour $a$ (remember that actions are simply encoded as the successor satellite), $L_{queue}(s, a)$ is the variable time that packets spend waiting in queues depending on the current congestion level, and $L_{target}(a) \in \{0, -100\}$ is a large negative reward that is only applied when the destination is reached by this action (and zero otherwise). This formulation encourages the agent to prioritize low-latency paths while penalizing queuing bottlenecks.

In our simulations, the queueing latency $L_{queue}(s, a)$ is not fixed but dynamically sampled at each step from a uniform distribution $\mathcal{U}(l_{\min}, l_{\max})$, individually configured for each satellite link. The bounds $l_{\min}$ and $l_{\max}$ are designed to reflect

**Algorithm 1** Pseudocode for the AI agent's training function.

1: **function** TRAINAGENT(*env*: SATELLITENETWORKENV, #*episodes*: $\mathbb{N}, \alpha: \mathbb{R}, \gamma: \mathbb{R}, \varepsilon: \mathbb{R}, epsilon\_decay: \mathbb{R}$)
2:    $Q \leftarrow \{0\}_{n \times n}$ ▷ initialize Q-table
3:    **for** *episode* = 1, ..., #*episodes* **do**
4:       *state* ← *env*.RESET() ▷ reset initial state to src
5:       *done* ← False
6:       *visited* ← {} ▷ track visited states in this episode
7:       **while** ¬*done* **do**
8:          *visited* ← *visited* ∪ {*state*} ▷ add state to visited
9:          *valid_actions* ← *env*.GETVALIDACTIONS(*state*) ▷ equivalent to $A^f(state)$
10:         **if** RANDOM(0, 1) < $\varepsilon$ **then** ▷ exploration-exploitation decision
11:            *action* ← RANDOMCHOICE(*valid_actions*) ▷ exploration: choose a random action
12:         **else**
13:            *action* ← $\arg\max_{va \in valid\_actions} Q(state, va)$ ▷ exploitation: choose the best action based on Q-table
14:         *new_state*, *reward*, *done* ← *env*.STEP(*action*)
15:         Q.UPDATE(*state*, *action*, *reward*, *new_state*, $\alpha, \gamma$) ▷ update Q-table (see Equation 2)
16:         **if** *new_state* ∈ *visited* **then** ▷ check for loops: end the episode and assign penalty
17:            *reward* ← *reward* − 100 ▷ optional: assign a large negative reward for looping behaviour
18:            Q.UPDATE(*state*, *action*, *reward*, *new_state*, $\alpha, \gamma$)
19:            **break**; ▷ end the episode
20:         *state* ← *new_state* ▷ move to new state
21:       $\varepsilon \leftarrow \varepsilon \cdot epsilon\_decay$ ▷ apply epsilon decay at the end of each episode
22:    **return** *Q*

**Algorithm 2** Pseudocode for the AI agent's step function.

1: **function** STEP(*env*: SATELLITENETWORKENV, *action*: A)
2:    *reward* ← ENV.GETREWARD(*state*, *action*) ▷ equivalent to R(*env.state*, *action*)
3:    *env.state* ← *action* ▷ actions are encoded as successor satellite states
4:    *done* ← *action* = *dst* ▷ done iff. destination state reached
5:    **return** *env.state*, *reward*, *done*

realistic fluctuations in queueing delay caused by varying network load and traffic conditions. This stochastic modelling allows the agent to learn routing strategies robust to congestion dynamics, as the latency samples directly shape the observed rewards over time. In the evaluation presented below, $l_{\min}$ and $l_{\max}$ are randomly assigned on a per-satellite basis. However, these parameters—or ideally, the empirical $\mathcal{U}(\cdot)$ distribution itself—could be derived from actual telemetry data collected from the operational constellation.

**Q-Routing.** We now assemble everything together and explain the details of the Q-routing approach. Q-routing is a Q-learning-based adaptive routing algorithm designed to manage packet routing [31]. The Q-learning algorithm [32] is based on a Bellman equation and defined by the following update rule:

$$Q(s, a) \leftarrow Q(s, a) + \alpha \left[ R(s, a) + \gamma \left( \max_{a' \in A^f(s')} Q(s', a') \right) - Q(s, a) \right] \quad (2)$$
$$= (1 - \alpha) \cdot Q(s, a) + \alpha \left[ R(s, a) + \gamma \left( \max_{a' \in A^f(s')} Q(s', a') \right) \right]$$

where $Q(s, a)$ is the Q-value for state $s$ and action $a$, $\alpha$ is the learning rate controlling how much new updates affect learning (e.g. $\alpha = 0.05$), $R(s, a)$ is the reward for taking action $a$ in state $s$, $\gamma$ is the discount factor weighting future rewards against present ones (e.g. $\gamma = 0.9$), and $\max_{a' \in A^f(s')} Q(s', a')$ is the maximum Q-value for the next state $s'$ (in this case equal to $a$).

The training phase of our model is given in Algorithm 1. Initially, we start with an empty Q-table. In each training episode, we first reset our environment so that the data packet is in state *src*. Then, in a loop, we execute actions (i.e. send the data packet from satellite to satellite) until the packet has reached the destination satellite *dst*. We use an $\varepsilon$-greedy approach (more precisely an $\varepsilon$-decreasing strategy) to select actions in order to balance between exploration and exploitation [28]. Initially, we set $\varepsilon = 0.3$ to discover new routes (line 11) and gradually decay $\varepsilon$ over time to favour exploitation of the learned values as the agent becomes more experienced (line 13). After each executed action, the Q-table is updated according to Equation 2 and returned by the function after all training episodes are done.

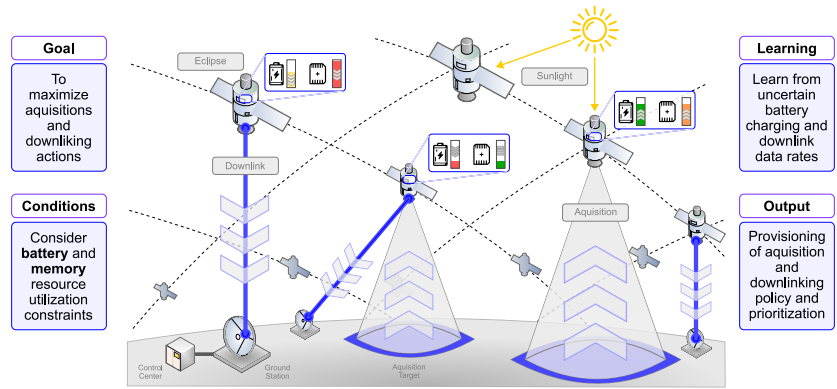

Fig. 3: Overview of Use Case 2: Resources.

The environment's step function (from line 14) is given in Algorithm 2. Its task is to compute the reward of the chosen action according to Equation 1, move the system's state to the successor state, and to check whether the current episode is done because the destination state has been reached.

## 4. Use Case 2: Resources

The second use case focuses on resource management within EO satellite constellations, where on-board resources such as battery power and memory are limited and must be managed efficiently to maximize overall productivity [33]. A sketch of this use case is given in Fig. 3. Here, we consider a set of EO satellites orbiting the Earth that are equipped with two typical EO payloads: (i) an acquisition payload (such as a camera or X-band SAR) used to collect data over certain areas of interest, and (ii) a transceiver (or transmitter) that can transfer payload data to a ground station. Since all satellites are following a pre-defined trajectory, the visibility episodes with ground stations and acquisition targets are known in advance.

The goal of our AI model is to maximize data acquisition and downlink activities while considering memory limitations and balancing energy consumption to prevent battery depletion in the future. Our RL agent learns from variable sunlight exposure for battery recharging and fluctuating data acquisition/downlink opportunities to come up with a near-optimal task schedule that optimizes resource utilization across the satellites.

### 4.1 Model Definition

In this scenario, we again consider a satellite constellation with $n$ satellites flying in pre-defined orbits, meaning that we can predict their future positions and compute access intervals in advance. The topology model is discretized into $T$ time slots of variable duration $\tau$. The variable slot durations $\tau_t$ are determined by changes in access to ground elements and sunlight exposure. This means that the time slots are chosen such that there are no changes to the scenario within each time slot $t \in \{0, 1, \ldots, T-1\}$. Along their trajectory, satellites encounter windows of opportunity to interact with (i) acquisition targets (ATs) for payload data generation, and (ii) ground stations (GSs) for payload data downlinking. Downlinking activities to GSs consume energy from the on-board battery, which is charged during sunlight exposure. Acquisition activities on ATs generate data stored in on-board memory, which is offloaded during downlinking. In our implementation, only downlink operations consume significant energy due to the power-intensive transmitter usage, while acquisitions are assumed to be energy-neutral for simplicity. This abstraction reflects scenarios where imaging sensors are low-power or where their energy cost is negligible relative to downlinking. Note that downlinks and acquisitions might occur simultaneously at a given satellite in a given time slot. However, without loss of generality, we assume that opportunity windows of the same type (i.e. GS or AT) are pairwise disjunct. In the following, we assume that there are $n$ satellites, $n_{gs}$ ground stations, and $n_{at}$ acquisition targets.

**State Space.** The state space $S$ (or observation space) represents the satellite constellation's current resource conditions and environmental context at each time slot. A state $s_t \in S$ at time slot index $t$ is modelled as a tuple $s_t = (\tau_t, sat_t^1, sat_t^2, \ldots, sat_t^n)$ where $\tau_t$ is the duration of the current time slot (in seconds) and $sat_t^i$ is the per-satellite observation for each satellite $i$. It is itself modelled as a tuple $sat_t^i = (bat_t, mem_t, opp_t^{at}, opp_t^{gs}, sun_t)$, where

(i) $bat_t \in [0, 1]$ is the (continuous, but discrete is possible) battery charge level of the satellite at the beginning of time slot $t$ (0.0 % ≜ depleted battery, 100.0 % ≜ full battery),

(ii) $mem_t \in [0, 1]$ is the (continuous, but discrete is possible) memory usage of the satellite at the beginning of time slot $t$ (0.0 % ≜ memory empty, 100.0 % ≜ memory full, no more storage available),

(iii) $opp_t^{at} \in \{0, 1, \ldots, n_{at}\}$ indicates the presence of an acquisition opportunity (0 ≜ no access possible, $opp_t^{at} = x > 0$ indicates access to AT with id $x$),

(iv) $opp_t^{gs} \in \{0, 1, \ldots, n_{gs}\}$ indicates the presence of a downlink opportunity (0 ≜ no access possible, $opp_t^{gs} = x > 0$ indicates access to GS with id $x$), and

(v) $sun_t \in \{0, 1\}$ is a binary indicator of whether the satellite is exposed to sunlight during time slot $t$ and can recharge its batteries (0 ≜ no sunlight, 1 ≜ exposed to sunlight).

**Action Space.** The action space consists of two types of tasks that each satellite can execute: (i) acquisition tasks $Q$ that are performed at areas of interest, i.e. ATs, and (ii) downlink tasks $D$ that are performed over a GS to offload collected data.

Since acquisitions and downlinks can also happen simultaneously, each satellite can select an action in each time slot $t$ from the following set of discrete actions $A_t = \{NOP, Q, D, QD\}$, where *NOP* means no action ("no-op", i.e. satellite is idle), $Q$ means to acquire payload data, $D$ means downlinking stored data to a ground station, and $QD$ means to perform both actions simultaneously.

Due to the orbital dynamics, not all actions are always feasible in every time slot. Therefore, the feasible action space $A_t^f$ at time $t$ is a subset of $A_t$ satisfying the following constraints: An acquisition action $Q$ requires an access opportunity to an acquisition target, i.e. $opp_t^{at} \neq 0$ for the respective satellite. Analogously, a downlink action $D$ requires a ground station link, i.e. $opp_t^{gs} \neq 0$. Acquisition and downlink actions are considered feasible but are heavily penalized if they exceed any battery or memory limits, e.g. deplete the battery or overflow the memory.

Finally, to accommodate the multi-satellite scenario with $n$ satellites, we lift the action space to encode the decision taken by each individual satellite and represent it as a tuple $(A_t^f)_1 \times (A_t^f)_2 \times \cdots \times (A_t^f)_n$.

**Transition Dynamics.** We now describe the effects of the different actions on the battery and memory levels. Downlink tasks consume energy while the battery is recharged during sunlight episodes. Acquisition tasks generate data that occupies on-board memory while the memory is freed up by downlink tasks. In addition, we consider a background load that always applies as well as a permanent increase in memory usage due to telemetry generation. All these effects happen proportional to the duration of the time slot, which means that we define constant rates of change for each parameter.

The battery and memory update functions for a chosen action $a_t \in A_t$ are therefore defined as follows:

$$bat_{t+1} \leftarrow bat_t + \tau_t \cdot \Big( \underbrace{\Delta B_{bg}^{\downarrow}}_{\text{background}} + \underbrace{\chi_{dl}(a_t, opp_t^{gs}) \cdot \Delta B_{dl}^{\downarrow}}_{\text{downlink}} + \underbrace{sun_t \cdot \Delta B_{sun}^{\uparrow}}_{\text{sun exposure}} \Big)$$

$$mem_{t+1} \leftarrow mem_t + \tau_t \cdot \Big( \underbrace{\Delta M_{tm}^{\uparrow}}_{\text{telemetry}} + \underbrace{\chi_{dl}(a_t, opp_t^{gs}) \cdot \Delta M_{dl}^{\downarrow}}_{\text{downlink}} + \underbrace{\chi_{aq}(a_t, opp_t^{at}) \cdot \Delta M_{aq}^{\uparrow}}_{\text{acquisition}} \Big)$$

We annotate each rate with an up-pointing arrow if it is positive (i.e. increases the respective quantity) and use a down-pointing arrow for negative rates. While the background load and telemetry generation always apply, the rate for sunlight exposure is only applied if the satellite is exposed to sunlight in the current slot (i.e. $sun_t = 1$). Moreover, the rates for a downlink or an acquisition are only applied if the chosen action $a_t$ performs the respective task and there exists an actual opportunity of that type, meaning that this task is feasible. This is ensured by the following binary indicator functions:

$$\chi_{dl}(a_t, opp_t^{gs}) := \begin{cases} 1 & \text{if } a_t \in \{D, QD\} \land opp_t^{gs} \neq 0 \\ 0 & \text{otherwise} \end{cases} \quad \text{(i.e. a \textit{feasible} downlink action)}$$

$$\chi_{aq}(a_t, opp_t^{at}) := \begin{cases} 1 & \text{if } a_t \in \{Q, QD\} \land opp_t^{at} \neq 0 \\ 0 & \text{otherwise} \end{cases} \quad \text{(i.e. a \textit{feasible} acquisition action)}$$

In addition, the battery and memory are limited to the interval [0, 1] because, for example, it is not possible to charge a battery to more than 100 % or to store more data than there is memory available. Preventing these situations is the job of the AI model where such behaviour is penalized by assigning negative rewards.

Optionally, our model supports to sample the battery and memory rates from continuous uniform distributions with given minimum and maximum parameters to simulate, for example, varying channel characteristics due to atmospheric effects or fluctuating solar incidence angles.

**Policy.** The policy maps the current state $s_t$ to a probability distribution over actions $A_t$. In our work, we use the Proximal Policy Optimization (PPO) algorithm [34], a simple yet efficient policy gradient method that encourages stable learning by preventing drastic policy shifts, which makes it well-suited for the complex nature of satellite resource management. PPO uses an advantage function to estimate the relative value of actions. To ensure stability, policy changes are limited between iterations by using a clip function. Further, we use a masked action model [35] so that the network is trained to avoid infeasible actions rather than relying on explicit constraints.

The policy itself is implemented as a Neural Network (NN) that receives the full constellation state $s_t$ as input and outputs a distribution over the feasible action space $A_t^f$. Note that by "feasible" actions, we refer to those permitted by the masking mechanism of the PPO model—specifically, acquisition tasks when the satellite is over an AT, and downlink tasks when over a GS. Conversely, actions that lead to battery depletion or memory overflow are still considered feasible in terms of action masking, but they are strongly penalized in the reward function to discourage resource-violating behaviour.

**Reward.** Rewards are shaped to maximize data throughput (i.e. collecting data and downlinking it to ground) while ensuring sustainable resource management. The reward function combines data acquisition and downlink rewards, together with penalties for battery depletion, memory overload, and conflicting antenna operations. The reward at time step $t$ for a single satellite and selected action $a_t$ is defined as:

$$R_t(a_t) := \begin{cases} \underbrace{\tau_t \cdot R_{aq}}_{acquisition} + \underbrace{\tau_t \cdot R_{dl}}_{downlink} - \underbrace{P_{bat}(bat_{t+1})}_{bat.\ depletion\ penalty} - \underbrace{P_{mem}(mem_{t+1})}_{mem.\ overflow\ penalty} - \underbrace{P_{simGS} - P_{simAT}}_{overlapping\ action\ penalties} & \text{if } a_t \text{ is feasible} \\ P_{invAct} & \text{otherwise} \end{cases}$$

where $R_{aq} := \chi_{aq}(a_t, opp_t^{at}) \cdot \Delta R_{aq}^{\uparrow}$ and $R_{dl} := \chi_{dl}(a_t, opp_t^{gs}) \cdot \Delta R_{dl}^{\uparrow}$ are the acquisition and downlink reward rates (that only apply if the respective task is possible and executed), $P_{bat}$ and $P_{mem}$ are optional penalties for battery depletion and memory overflow, and $P_{simGS}$ and $P_{simAT}$ are optional penalties for conflicting acquisition or downlink actions. The penalty terms $P_{simGS}$ and $P_{simAT}$ are designed to discourage the simultaneous use of the same AT or GS by multiple satellites within the same time slot, which would lead to access contention and reduced performance. The current model adopts a simple greedy assignment strategy based on a first-come, first-served policy. Although this method does not guarantee an optimal matching between satellites and ground assets, it is straightforward to implement within the Gymnasium environment. Future work will explore embedding optimal assignment mechanisms directly into the environment. Nevertheless, the imposed penalties effectively guide the agent to stagger access over time to deliver a realistic resource allocation policy.

Thus, for the multi-satellite case with $n$ satellites, the total reward is expressed as $R_t = \sum_{i=1}^{n}(R_t)_i$, where $(R_t)_i$ is the reward for each satellite $i$.

## 5. Evaluation

We now present selected results from our extensive verification and validation analysis. For both use cases, we compare the performance of the RL-based algorithms with traditional baselines. We also investigate how the choice of learning parameters affects the quality of the resulting solutions. Finally, we perform a flexibility analysis to assess how the algorithms perform in failure scenarios.

### 5.1 Use Case 1: Routing

The evaluation of the routing use case compares Q-routing, our Reinforcement Learning approach, with two baselines, Dijkstra and Dijkstra MQ, to optimize data packet forwarding in satellite networks. The analysis focuses on minimizing end-to-end latency while considering dynamic queueing conditions and network congestion. The study assesses the adaptability of RL-based routing compared to deterministic shortest-path algorithms, highlighting trade-offs in computational efficiency, flexibility, and performance stability.

The (plain) Dijkstra baseline is a classical shortest-path algorithm without any traffic awareness. It uses only the (static) propagation latency as the optimization metric and does take into account queueing delays at intermediate hops.

The second baseline, Dijkstra MQ, is a variant of Dijkstra that has privileged access to (mean) queueing delay information from intermediate nodes. Here, the algorithm is aware of all network conditions in advance, which corresponds to a theoretical optimum (similar to a "God mode") and serves as a benchmark for comparing the effectiveness of the Q-routing approach in real-world scenarios.

*5.1.1 End-to-End Latency Analysis*

Fig. 4 presents the end-to-end latency distributions of Q-routing, Dijkstra, and Dijkstra MQ across different training episodes and learning rates, highlighting their respective performance characteristics.

Dijkstra MQ consistently achieves the lowest latencies due to its ability to account for queueing delays in real-time. In contrast, Dijkstra, though stable and deterministic, does not adapt to queueing dynamics and thus experiences higher latencies under congestion. Q-routing, the RL-based approach, initially exhibits high variance in its results.

The impact of hyperparameters on Q-routing is evident in the figure. As the number of training episodes increases (from $E = 15\,000$ to $E = 25\,000$), the latency distributions narrow, indicating improved routing stability. However, Q-routing stochastically presents more latency outliers, often corresponding to paths with higher hop counts. This suggests that while Q-routing can optimize average latency over time, it may still select suboptimal routes in some instances, mainly if training is insufficient.

The learning rate $\alpha$ also plays a crucial role. A low learning rate (e.g. $\alpha = 0.01$) ensures gradual but stable convergence, while a high learning rate (e.g. $\alpha = 0.1$) accelerates adaptation at the cost of increased variability. The results indicate that a moderate learning rate ($\alpha = 0.05$) offers the best trade-off, balancing convergence speed and routing stability.

*5.1.2 Latency Gain Analysis*

Fig. 5 presents a detailed latency comparison for scenarios where Q-routing achieves better performance than Dijkstra. The x-axis lists different source-target node pairs and their average hop counts, while the y-axis represents the average end-to-end latency in milliseconds. We distinguish between the three algorithms: Q-routing (blue), Dijkstra (red), and Dijkstra MQ (green).

The plot shows that the performance of Q-routing can surpass that of Dijkstra across multiple node pairs. Above the bars, a label quantifies the latency gain of Q-routing relative to Dijkstra, measured as the difference in average end-to-end latency between the two algorithms. Gains range from fractions of a millisecond up to several milliseconds, indicating that when adequately trained, Q-routing can dynamically optimize routing decisions to reduce latency. While Q-routing improves over Dijkstra, Dijkstra MQ often achieves similar or even better latency, particularly on longer routes. This suggests that while Q-routing is effective in learning queue-aware routing policies, it does not always reach the efficiency of a theoretical optimum like Dijkstra MQ, which is fully aware of queueing conditions.

*5.1.3 Latency vs. Hop Count Analysis*

Fig. 6 illustrates the relationship between the latency difference and the hop count difference between Q-routing and Dijkstra for three training episode counts: 15 000, 20 000, and 25 000. The x-axis measures the latency difference, which is calculated as Q-routing's end-to-end latency minus that of Dijkstra. In contrast, the y-axis measures the difference in hop count, representing Q-routing's additional hops relative to Dijkstra.

Each data point corresponds to a unique source-target pair in the network. The red dashed trend line fitted to the data in each subplot provides a linear regression analysis of this relationship. The positive slope across all training scenarios indicates that more considerable latency differences often coincide with more significant hop count differences. The trend lines remain consistent across training levels, reinforcing that the fundamental relationship between hop count and latency difference remains stable, even as absolute deviations shrink. While additional training episodes improve Q-routing's efficiency, the results highlight a key limitation: the inherent stochasticity in learned policies can still lead to suboptimal path selections for some network conditions.

*5.1.4 Flexibility Analysis*

Fig. 7 examines the adaptability of Q-routing compared to traditional deterministic approaches like Dijkstra and Dijkstra MQ under failure scenarios. The three subplots analyse end-to-end latency, average hops, and average reward as failure size functions, representing the number of unresponsive nodes encountered along the computed path.

Unlike Dijkstra-based methods, which pre-compute fixed paths based on static network conditions, Q-routing dynamically adjusts to network failures by leveraging Reinforcement Learning. Its Q-table retains multiple viable paths, allowing it to adapt in real-time. The first plot on end-to-end latency illustrates this flexibility: while Dijkstra and

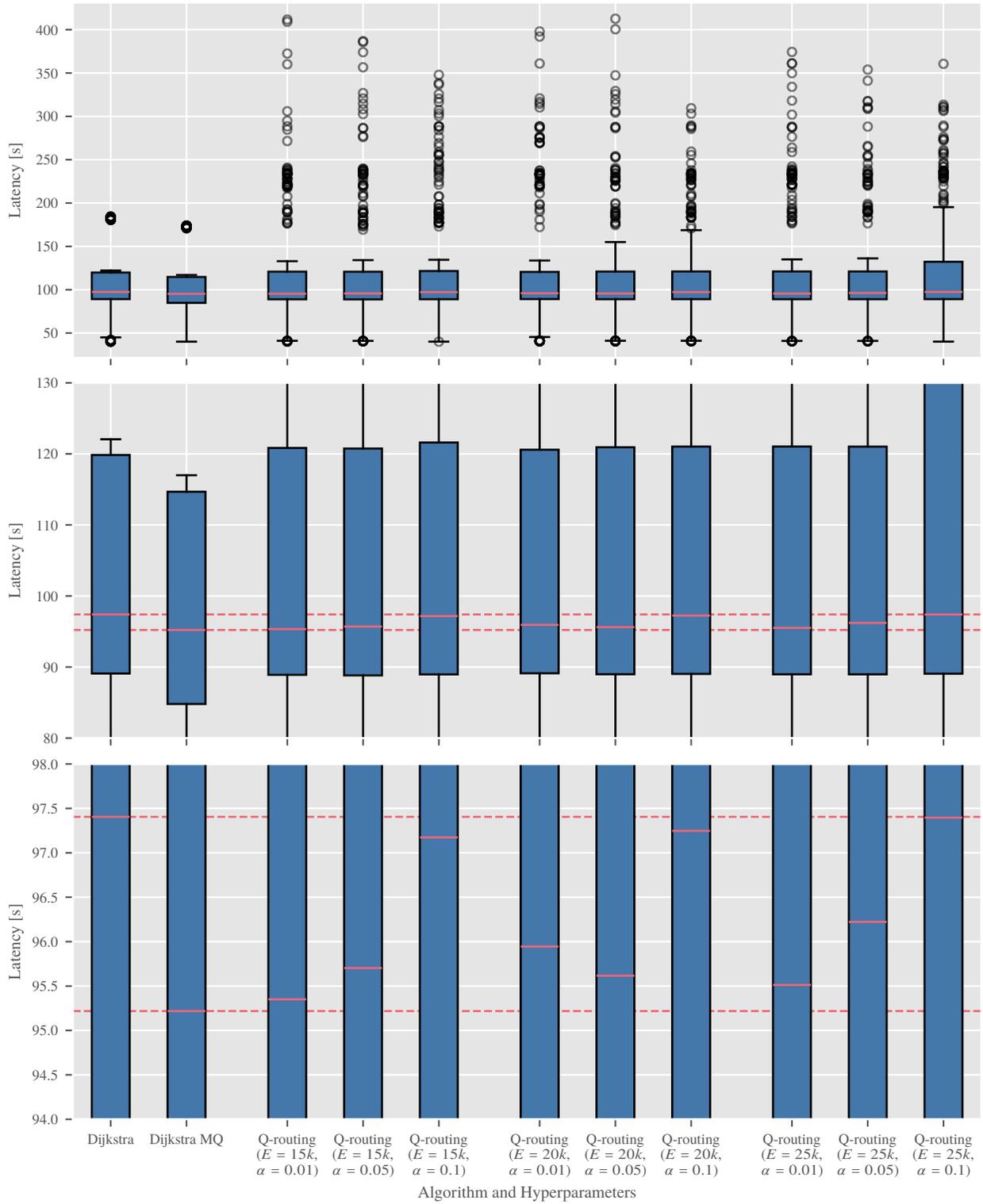

Fig. 4: End-to-end latency distribution for Q-routing variants and deterministic algorithms by zoom levels.

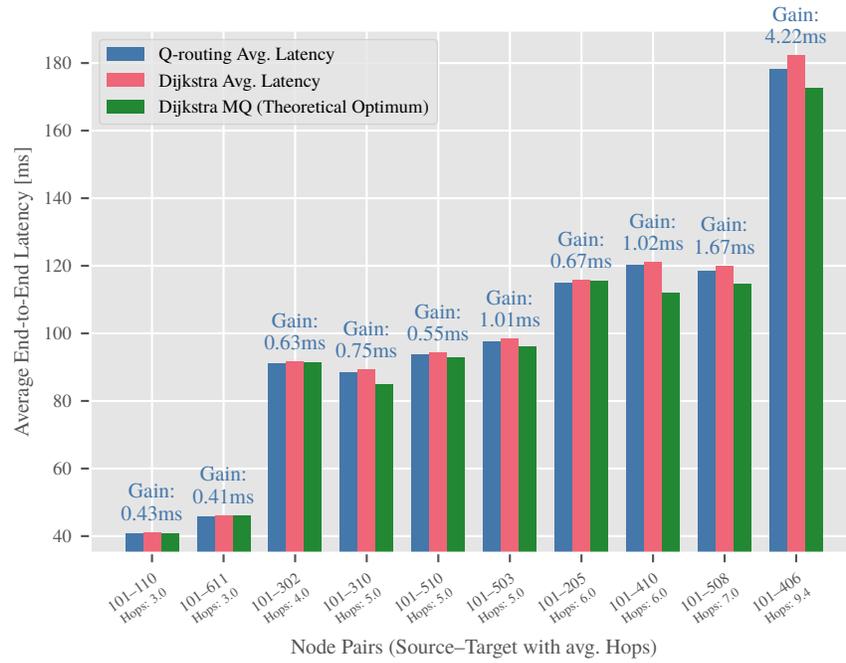

Fig. 5: Latency comparison for cases where Q-routing performs better than Dijkstra.

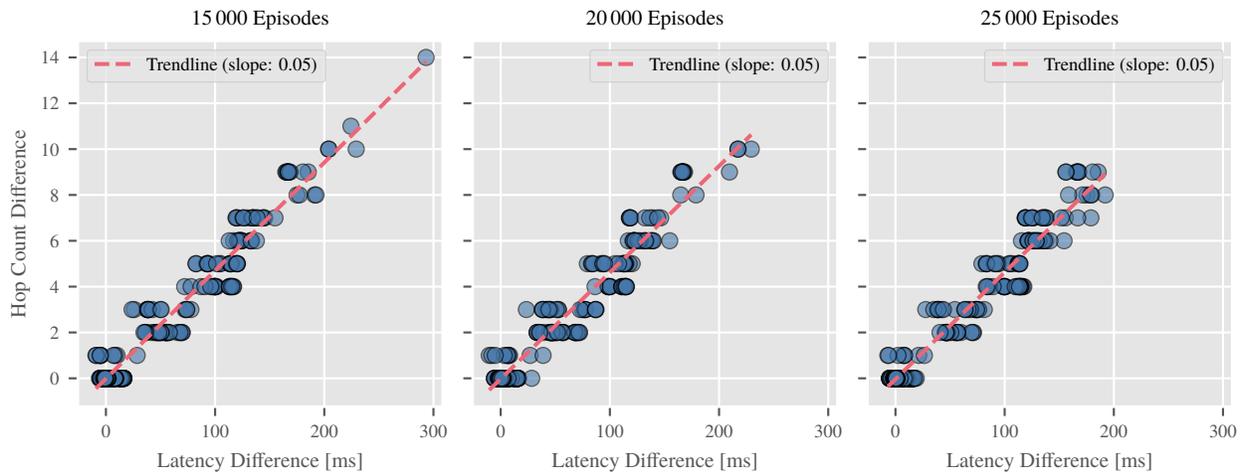

Fig. 6: Latency difference vs. hop count difference for different episode counts.

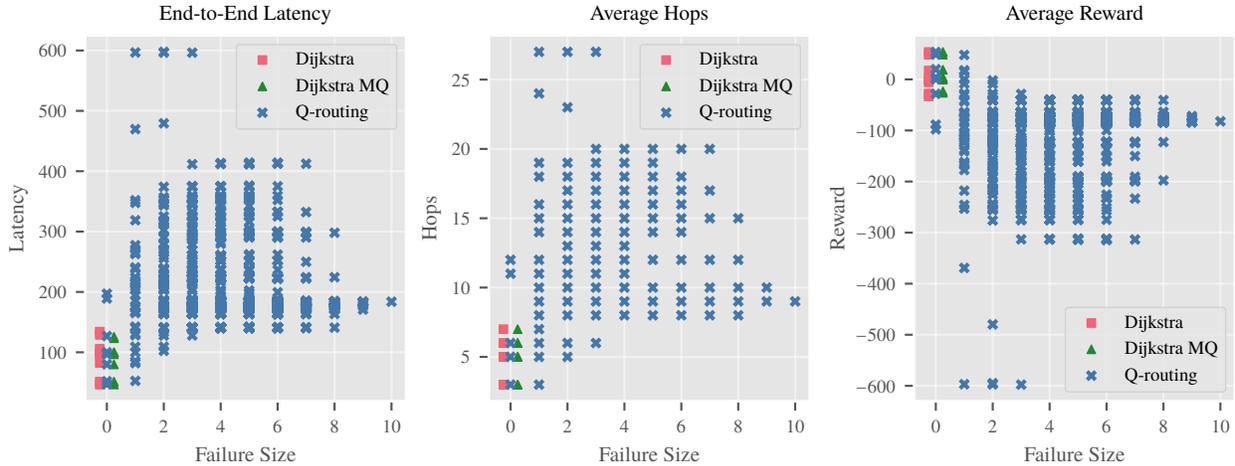

Fig. 7: Flexibility analysis. The failure size is given in terms of the number of unresponsive nodes in the path.

Dijkstra MQ (represented by red squares and green triangles) maintain optimal latency in failure-free conditions, they fail once disruption occurs. In contrast, Q-routing (blue crosses) continues to find alternative paths, albeit at the cost of increased latency as failure grows, reflecting longer detours.

The second plot, average hops, follows a similar trend. The third plot, average reward, highlights the efficiency trade-offs. While deterministic approaches maintain high reward values in stable conditions, their reward drops to zero when failures occur, indicating an inability to recover. On the other hand, Q-routing shows a steady decline in reward, reflecting increased path costs but retaining connectivity in all cases.

Even under failure conditions, this ability to adapt makes Q-routing a robust choice for networks with frequent and unpredictable link failures. Ultimately, Q-routing sacrifices some optimality in latency and hop count but ensures continued network operation where deterministic methods fail. This makes it particularly suitable for mission-critical applications where packet delivery is more important than strict efficiency.

### 5.1.5 Discussion and Key Takeaways

When sufficiently trained, Q-routing achieves latency reductions of a few milliseconds over Dijkstra in dynamic queueing conditions. However, this improvement comes at a cost—Q-routing demands extensive training to converge, increasing processing overhead. In contrast, deterministic algorithms like Dijkstra and Dijkstra MQ, though lacking real-time adaptability, offer stable, computationally efficient solutions that remain practical for many scenarios.

Q-routing's performance improves as training episodes increase, leading to lower latency and a more stable distribution. However, higher hop counts correlate with more significant latency variations, indicating that additional training is needed to minimize these effects. While Q-routing can surpass Dijkstra in select cases, its stochastic nature introduces inconsistencies, particularly over longer paths. Thus, its overall performance is less predictable without extensive fine-tuning.

A key takeaway from the flexibility analysis is Q-routing's ability to sustain connectivity in failure-prone networks. Unlike Dijkstra and Dijkstra MQ, which fail when their pre-computed paths become unavailable, Q-routing dynamically selects alternative routes from its learned Q-table. This enables continuous packet delivery, though at the expense of increased latency and hop count as failures grow. While deterministic methods can support multiple paths through explicit provisioning (e.g. K-best routing [36, 37]), this adds operational overhead, making Q-routing a compelling alternative for environments with unpredictable failures.

### 5.2 Use Case 2: Resources

The evaluation of the resource management use case compares Reinforcement Learning (RL), Simulated Annealing (SA), and a Randomized (RND) heuristic to optimize satellite resource allocation while accounting for constraints such as battery depletion, memory overflow, and scheduling conflicts.

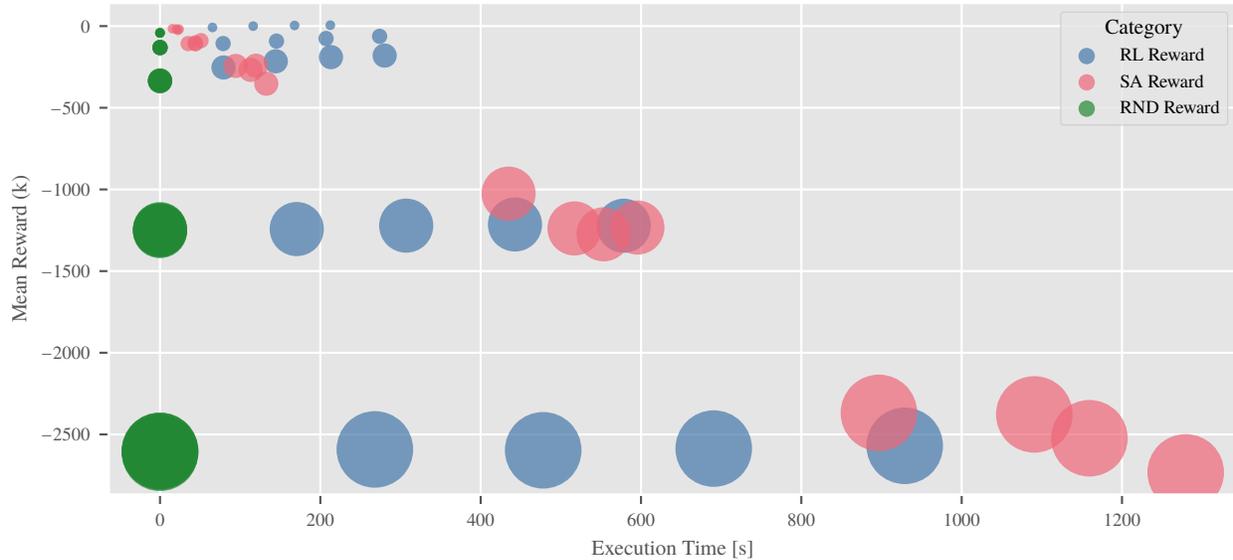

Fig. 8: Execution time vs. mean reward for the Resources use case.

### 5.2.1 Reward Function and Execution Time Trade-Off Analysis

The reward function integrates multiple objectives: data acquisition and downlink rewards, along with penalties for battery depletion, memory overload, and conflicting antenna operations. These constraints ensure that the optimization strategies maximize data throughput while maintaining resource efficiency and operational feasibility.

Experiments were conducted across different satellite configurations, ranging from 10 to 100 satellites, and under varying training conditions. RL models were trained over 5, 10, 15, and 20 episodes, while SA was tested with 30, 60, 90, and 120 iterations. The marker sizes in Fig. 8 reflect scenario complexity, with larger markers representing more satellites and time slots.

The figure illustrates the trade-off between execution time and reward across the three approaches. The x-axis represents execution time (in seconds), while the y-axis indicates the mean reward (in thousands).

- Reinforcement Learning (RL) in blue: Achieves high rewards but requires significantly longer execution times, making it computationally demanding for large-scale problems.

- Simulated Annealing (SA) in red: Provides a balanced approach, offering competitive rewards with lower execution times than RL, particularly in complex scenarios.

- Randomized (RND) in green: Computationally inexpensive but consistently underperforms in reward optimization.

In simpler scenarios, RL and SA achieve comparable results. However, as scenario complexity increases, SA consistently outperforms RL in absolute reward, maintaining an edge in large-scale cases. The largest markers, corresponding to 50 and 100 satellites, highlight this widening performance gap, demonstrating SA's ability to scale more efficiently than RL.

### 5.2.2 Reward Deviation Analysis

Fig. 9 presents the average standard deviation of rewards for RL, SA, and RND approaches, highlighting the stability of each optimization method.

SA exhibits the lowest standard deviation, confirming its ability to produce consistent and predictable results across different runs. This makes SA particularly well-suited for complex scenarios where maintaining reliable performance is essential. On the other hand, RL shows moderate variability, indicating that performance fluctuates depending on training conditions and network states. While RL can achieve high rewards, its reliance on learned policies introduces inherent unpredictability, making its performance less stable than SA. Finally, RND displays the highest standard deviation, reinforcing its lack of structured optimization. Its inconsistency highlights the limitations of a purely random search approach, making it the least reliable in terms of stability.

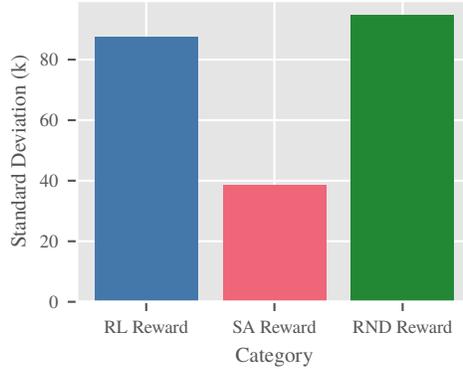

Fig. 9: Average standard deviation for the RL, SA, and RND approaches.

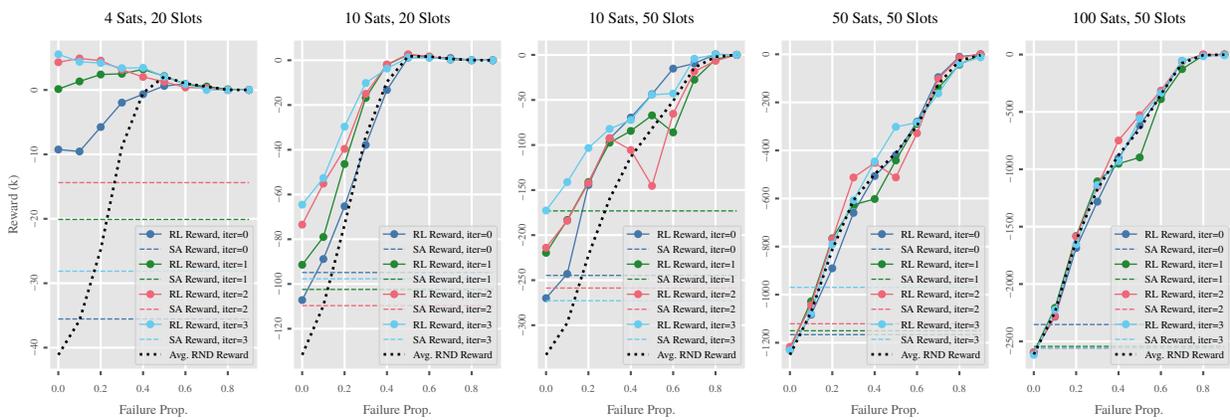

Fig. 10: Reward obtained for diverse failure proportion across scenarios.

These results emphasize that SA offers the most stable outcomes, making it the preferred choice when predictability is a priority. While adaptable, RL remains a viable option only when the benefits of learning-driven optimization outweigh the risk of performance fluctuations.

*5.2.3 Flexibility Analysis*

The flexibility evaluation of the resource management use case assesses the adaptability of RL and SA under unforeseen failures in access windows. The x-axis in Fig. 10 represents the failure proportion, while the y-axis indicates the reward, with higher values signifying better scheduling efficiency. The goal is to determine how well each approach performs when the network topology deviates from initial training conditions.

Five test cases are analysed, ranging from a 4-satellite, 20-slot scenario to a 100-satellite, 50-slot scenario. RL is trained for 5, 10, 15, and 20 episodes, while SA is evaluated with 30, 60, 90, and 120 iterations. The RND baseline is included (black dashed line) as a reference for unoptimized scheduling.

In simpler cases, such as 4 satellites and 20 slots, RL exhibits strong adaptability. Even as the failure proportion increases, RL maintains higher rewards than SA, particularly with sufficient training. RL's ability to dynamically adjust policies allows it to outperform the RND baseline, sustaining reasonable performance up to 50 % failure proportion. In contrast, SA remains static, producing a fixed schedule optimized for a failure-free scenario. Once deviations occur, SA's performance degrades, as it cannot re-optimize in real-time.

As scenario complexity grows (10 satellites, 50 slots), RL's advantage diminishes. At failure proportions above 20 %, RL's performance begins to converge with the RND heuristic, indicating that its ability to adapt weakens under increasingly unpredictable conditions. Meanwhile, SA, though still inflexible, matches RL's performance at low failure rates, suggesting that its deterministic scheduling remains competitive in moderately stable environments.

RL's flexibility advantage is significantly reduced in larger-scale cases (50+ satellites, 50 slots). With 50 satellites,

RL training plateaus, meaning additional iterations do not yield noticeable improvements. Despite its lack of adaptability, SA achieves competitive performance by leveraging its optimized scheduling approach under non-failure conditions.

*5.2.4 Discussion and Key Takeaways*

The evaluation of the resource management use case underscores a fundamental trade-off between RL and SA in terms of optimization performance, stability, and adaptability. RL excels in dynamically adjusting to network disruptions, making it highly effective in smaller-scale scenarios where real-time adaptability is crucial. However, as network complexity increases, SA consistently outperforms RL in absolute reward and stability, offering a more computationally efficient solution.

A key takeaway from the reward function analysis is that RL can achieve competitive performance only with extensive training. While SA remains inflexible—providing optimized schedules tailored for non-failure conditions—it maintains high performance with minimal variance, making it the more reliable choice for large-scale satellite networks. The reward deviation results further support this, with SA exhibiting the lowest standard deviation, whereas RL's performance fluctuates based on training conditions. The RND heuristic remains the weakest approach, lacking structured optimization and exhibiting high variability.

The flexibility analysis further clarifies RL's strengths and weaknesses. RL maintains a clear advantage over SA for low to moderate failure rates, particularly in smaller networks where deviations from training conditions are minimal. However, as failure proportions increase, RL's performance deteriorates, gradually converging with the RND baseline, indicating a loss of optimization capability under extreme disruptions. In contrast, SA remains rigid but consistently reliable, offering stable performance without additional adaptation.

At higher network complexities (50+ satellites), RL's computational overhead outweighs its benefits—additional training iterations do not significantly improve performance. While RL remains a viable solution for low-complexity and dynamically evolving environments, its limitations become evident in large-scale networks where training constraints and high failure rates result in diminishing returns. Conversely, SA proves to be a scalable and computationally efficient alternative, maintaining strong performance even as system complexity grows.

Ultimately, the choice between RL and SA depends on the system's operational needs. RL is preferable for small- to medium-scale networks where adaptability is essential and failure rates are moderate. However, in large-scale environments, SA emerges as the more efficient and stable approach, achieving higher rewards with lower variability and significantly reduced training costs.

## 6. Technical Infrastructure

Before we conclude the paper, we give a short overview of the technical infrastructure of our toolchain. In both use cases, the core logic of the AI engine and simulation environment is implemented in the Python programming language. The satellite network environment is modelled using the Gymnasium library [38], the de facto API standard for Reinforcement Learning. In the routing use case, we implement the code for Q-routing without relying on external libraries due to its conceptual simplicity. For the resources use case, we leverage the RLlib library [39] which provides a production-level implementation of PPO, and TensorFlow [40] to represent the internal Neural Network.

The simulation environment and the AI engine are embedded in a containerized infrastructure. User interaction is provided through a ReactJS web front end and controlled through a FastAPI back end. The back end configures the simulator tool and enables the data preparation between the simulator and the AI engine. The results are distributed to the front end for visual presentation and user downloads in image and text formats. All components are orchestrated with Docker Compose, which allows a consistent and reproducible deployment containing all required dependencies and version consistency.

## 7. Conclusions

In summary, the rapid growth of satellite constellations requires a shift from traditional, human-centric operations to AI-driven optimization techniques. This research demonstrates that RL is competitive with classical approaches and offers enhanced flexibility, scalability, and generalizability in decision-making processes, which is critical for the autonomous and intelligent management of satellite fleets. The implications of this activity suggest that AI can fundamentally alter the landscape of satellite constellation management by providing more adaptive, robust, and cost-effective solutions.

In addition, our findings have highlighted weaknesses of centralized AI approaches. This encourages continued research of solutions where the AI training and/or inference is performed onboard the satellites—which was beyond the scope of this activity—with the obvious drawback of more severe on-board computational constraints. Future work should further explore other operational challenges where AI and RL can potentially be applied. For example, autonomous collision avoidance & space traffic management, or fault detection & autonomous anomaly handling, could both benefit from AI-enabled solutions to eliminate the response time delays associated with traditional ground-based monitoring.

**Acknowledgements**

This project has received funding from the European Space Agency's E/0904-611 – GSTP Element 1: Develop programme under ESA Contract No 4000141691/23/D/SR. The authors would like to thank all colleagues from ESA and commercial operators for their contributions to the ConstellAI project and invaluable feedback to the developments.